\documentclass[twoside,leqno,twocolumn]{article}

\usepackage[letterpaper]{geometry}

\usepackage{ltexpprt}
\usepackage{graphicx}
\usepackage{amsmath,bm}
\usepackage{hyperref}
\usepackage{color}
\usepackage{amssymb}

\providecommand{\keywords}[1]
{
  \small	
  \noindent
  \textbf{Keywords:\\ }  
}

\begin{document}

\title{WaveletAE: A Wavelet-enhanced Autoencoder for Wind Turbine Blade Icing Detection}

\author{Binhang Yuan\thanks{Rice University, \{by8,cl67\}@rice.edu.}
\and Chen Wang\thanks{Tsinghua University, \{wang\_chen,mingsheng\}@tsinghua.edu.cn.}
\and Chen Luo\footnotemark[1]
\and Fei Jiang\thanks{The University of Hong Kong, feijiang@hku.hk.}
\and Mingsheng Long\footnotemark[2]
\and Philip S. Yu\thanks{University of Illinois at Chicago, Philip S. Yu is also with the Institute for Data Science, Tsinghua University, psyu@cs.uic.edu.}
\and Yuan Liu\thanks{Goldwind. Inc, liuyuan@goldwind.com.cn.}}
\date{}

\maketitle


\fancyfoot[R]{\scriptsize{Copyright \textcopyright\ 2020 by SIAM\\
Unauthorized reproduction of this article is prohibited}}





\begin{abstract} \small
Wind power, as an alternative to burning fossil fuels, is abundant and inexhaustible. 
To fully utilize wind power, wind farms are usually located in areas of high altitude and facing serious ice conditions, which can lead to serious consequences.
Quick detection of blade ice accretion is crucial for the maintenance of wind farms. 
Unlike traditional methods of installing expensive physical detectors on wind blades, data-driven approaches are increasingly popular for inspecting the wind turbine failures.
In this work, we propose a wavelet enhanced autoencoder model (WaveletAE) to identify wind turbine dysfunction by analyzing the multivariate time series monitored by the SCADA system.
WaveletAE is enhanced with wavelet detail coefficients to enforce the autoencoder to capture information from multiple scales, and the CNN-LSTM architecture is applied to learn channel-wise and temporal-wise relations. The empirical study shows that the proposed model outperforms other state-of-the-art time series anomaly detection methods for real-world blade icing detection.

\end{abstract}


\section{Introduction}
\label{Introduction}

Worldwide, nearly a billion people lack access to electricity, and around 3 billion people rely on dirty fuels, such as wood and animal dung, for cooking.
As an alternative to burning fossil fuels, wind power is an important sustainable energy. 
To utilize wind power, wind turbines are applied to capture kinetic energy from the wind and generate electricity.
Wind energy is clean, renewable, and widely distributed, which makes it one of the fastest-growing energy sources in the world \cite{fthenakis2009land}. 

However, to get enough wind for electronic generation, most wind farms are located in areas of high-altitude with a high probability of ice occurrence.
Icing conditions pose a serious challenge to turbine blades. 
According to the statistics from wind farms built by Goldwind Inc \footnote{Goldwind Inc is the largest wind turbine manufacturer in China, and the third largest in the world of 2018. Goldwind has installed a total capacity of 41GW wind turbines in over 20 major countries around the world \cite{goldwind}.}, there are up to $25\%$ annual energy production losses due to blade icing. 
Ice accumulated on a blade will typically cause degradation of a turbine’s aerodynamic performance, and cause many other serious problems, such as measurement errors, overproduction, mechanical failures, and electrical failures \cite{parent2011anti}. 
To minimize losses caused by icing conditions, much effort has been invested in early icing condition detection.

\begin{figure}[t]
	\centering
	\includegraphics[width=0.45\textwidth]{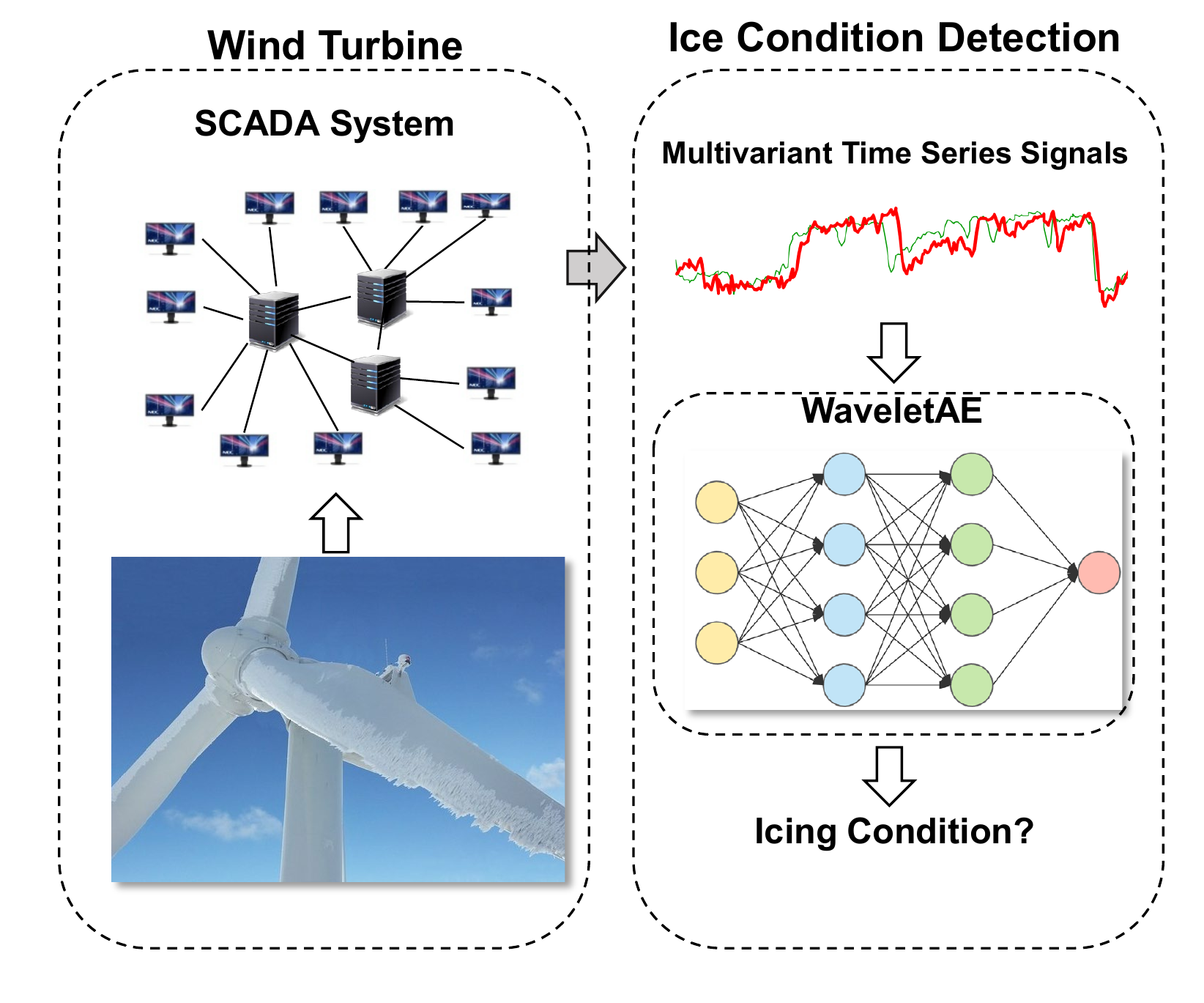}
	\caption{Using WaveletAE to detect ice condition by analyzing multivariate time series signals collected by the SCADA system on wind turbines.}
	\label{figure_frame}
\end{figure}

Traditionally, applied physics and mechanical engineering researches try to resolve this problem by designing and installing new physical detectors. 
Various techniques, such as damping of ultrasonic waves \cite{luukkala1995detector}, measurement of the resonance frequency \cite{carlsson2010measuring}, thermal infrared radiometry \cite{munoz2016ice}, optical measurement techniques \cite{pegau1992optical}, ultrasonic guides waves \cite{munoz2018wavelet} and etc., have been applied for icing detection. 
However, these techniques are limited by high costs and energy consumption. 
Besides, the ice sensors may provide inaccurate estimates of icing risks for wind turbines due to the internal unreliability \cite{parent2011anti}. 
Worse still, the installation of such detectors may require some unstable mechanical change of the wind turbine, and demand huge manpower to place the sensors in the turbine nacelle and blades. 


To reduce the risk and expense of detector installation and avoid the mechanical change of the turbine, in this work, we propose a data-driven approach to analyze the signals from the standard pre-installed sensors in the wind turbine in an attempt to design a deployable model for blade icing detection. In practice, the supervisory control and data acquisition (SCADA) system manages the installed general-purpose sensors monitoring the weather and turbine conditions, such as wind speed, internal temperature, yaw positions, pitch angles, power output, etc. Figure \ref{figure_frame} illustrates the workflow of the proposed data driven approach for icing detection. 

On the other hand, there are some fundamental challenges to analyze the multivariate signals by the classic time series anomaly detection algorithms: 
\begin{itemize}
	\item The multivariate signals usually include complicated correlation among channels, and learning the knowledge from such correlation is the key to identify blade icing dysfunction. For example, the power output is usually determined by wind speed and direction, and the corresponding working states are reflected in signals from multiple pitch angle, speed, and direct current sensors; the iced blade will arouse implicit change of the correlation among the signals.
	
	\item Ice accretion may show different dynamics due to various weather conditions. The reflected signal change is highly depended on the frequency, duration, severity, and intensity of icing. As a result, the proposed model should be able to detect changes of the pattern embedded in both time-domain and frequency-domain of the multivariate signals. 
\end{itemize}

To address the above challenges, we propose WaveletAE, a wavelet enhanced autoencoder model. 
WaveletAE first augments the original signal with wavelet detail coefficients, which reveal the signal variance in multiple scales to disclose the information in both time and frequency domains. 
Then, in each scale, the multivariate signals first go through a convolutional encoder to learn the global correlations among all the signal channels.
After convolutional encoder, the output goes through an LSTM encoder to capture the temporal complex, storing the information in the hidden states.
Once the LSTM encoder visits the whole multivariate signals, the final hidden states from all scales will be concatenated to generate the global hidden state.
Symmetrically, in the decoding phase, the final hidden states in each scale will be initialized by a mapping of the global hidden state according to independent fully connected layers.
Then, the hidden states will go through the LSTM decoder in reverse order followed by the de-convolutional layer to reconstruct the wavelet detail coefficients and the original signals. 
The main contributions of this work are listed below:

\begin{itemize}
	\item We propose WaveletAE, a generative autoencoder architecture that includes discrete wavelet transform to encode the multi-scale decomposition information for time series anomaly detection.
	
	\item WaveletAE is able to capture the complicated correlations and various dynamics in both frequent and temporal domains, under both semi-supervised and supervised settings.
	
	\item The experimental result demonstrates the effectiveness of WaveletAE on real-world data sets. Besides, a case study of simulated deployment suggests the robustness and flexibility of WaveletAE in real-time monitoring.
	

\end{itemize}

\section{Preliminary}
\label{preliminary}
We begin the introduction of our approach by providing a brief review of some necessary techniques: the discrete wavelet transform, and the deep autoencoder architecture. Then, we formally define the problem of anomaly detection in the blade icing accretion scenario. 

\subsection{Discrete Wavelet Transform}
The discrete wavelet transform decomposes a discrete time signal into a discrete wavelet representation \cite{chun2010tutorial}. Formally, given
$\boldsymbol{x} = {\left[ {\begin{array}{*{20}{c}}
{{x_0}}&{{x_1}}&{...}&{{x_{N - 1}}}
\end{array}} \right]^T}$ 
that represents a length N signal, and the basis functions of the form
$\boldsymbol{\varphi}  = {\left[ {\begin{array}{*{20}{c}}
{{\varphi _0}}&{{\varphi _1}}&{...}&{{\varphi _{N - 1}}}
\end{array}} \right]^T}$
and
$\boldsymbol{\psi}  = {\left[ {\begin{array}{*{20}{c}}
{{\psi _0}}&{{\psi _1}}&{...}&{{\psi _{N - 1}}}
\end{array}} \right]^T}$, 
then the coefficients for each translation (indexed by $k$) in each scale level (indexed by $j_0$ or $j$) are projections of the signal onto each of the basis functions:
\begin{equation}
\begin{array}{l}
{\boldsymbol{w}_\varphi }\left[ {{j_0},k} \right] = \left( {\begin{array}{*{20}{c}}
{\boldsymbol{x},}&{{\varphi _{{j_0},k}}}
\end{array}} \right) = \frac{1}{{\sqrt N }}\sum\limits_{m = 0}^{N - 1} {\boldsymbol{x}\left[ m \right]{\varphi _{{j_0},k}}\left[ m \right]} \\
{\boldsymbol{w}_\psi }\left[ {j,k} \right] = \left( {\begin{array}{*{20}{c}}
{\boldsymbol{x},}&{{\psi _{j,k}}}
\end{array}} \right) = \frac{1}{{\sqrt N }}\sum\limits_{m = 0}^{N - 1} {\boldsymbol{x}\left[ m \right]{\psi _{j,k}}\left[ m \right]} 
\end{array}
\end{equation}
where ${\boldsymbol{w}_\varphi }\left[ {{j_0},k} \right]$ is called approximation coefficient, and ${\boldsymbol{w}_\psi }\left[ {j,k} \right]$ is called detail coefficient.

The detail coefficients at different levels reveal variances of the signal on different scales, while the approximation coefficient yields the smoothed average on that scale. One important property of the discrete wavelet transform is that detail coefficients at each level are orthogonal, that says for any pair of detail coefficients not in the same level, the inner product is $0$:
\begin{equation}
{\boldsymbol{w}_\psi }\left[ {j, * } \right] \cdot {\boldsymbol{w}_\psi }\left[ {j', * } \right] = 0
\end{equation}
As a result, we can interpret the detail coefficients as an additive decomposition of the signal known as multi-resolution analysis. 

\subsection{Deep Autoencoder}
A deep autoencoder is a multi-layer neural network, in which the desired output is the original input. 
Internally, the autoencoder includes a hidden layer $\boldsymbol{h}$ that describes a low-dimensional code to represent the input $\boldsymbol{x}$. The network consists of two parts: an encoder function $\boldsymbol{h} = f_{E}\left( \boldsymbol{x} \right)$ parameterized by $E$ that maps the input signal $\boldsymbol{x}$ to a code and a decoder function $\boldsymbol{\bar{x}} = f_{D}\left( \boldsymbol{h} \right)$ parameterized by $D$ that produces a reconstruction $\boldsymbol{\bar{x}}$ of the input. Intuitively, learning a low dimensional representation forces the autoencoder to capture the most salient features of the training data.

The learning process is defined simply by minimizing a $loss$ function:

\begin{equation}
\min_{E,D} loss \left(\boldsymbol{x}, \: f_{D}\left(f_{E}\left(\boldsymbol{x}\right)\right)\right)
\end{equation}

where the $loss$ function penalizing $f_{D}\left(f_{E}\left(\boldsymbol{x}\right)\right)$ for being dissimilar from $\boldsymbol{x}$, commonly chosen to be the mean squared error. Usually, the encoder and decoder share symmetric architectures by stacking many neural network layers in reverse order.

\subsection{Problem Formalization}
\label{preliminary_3}
Time series anomaly detection has been formalized in multiple ways in different applications. 
Formally, we define the anomaly detection in this work as below for the blade icing accumulation scenario.

Given the multivariate time series $\boldsymbol{x} \in R^{C\times T}$, where $C$ is the number of channels, and $T$ is the length of the signal, denoted equivalently by

\begin{itemize}
	\item $\boldsymbol{x} = \left[ \boldsymbol{x}^{(0)}, \boldsymbol{x}^{(1)}, ..., \boldsymbol{x}^{(T)}\right]$ where $\boldsymbol{x}^{(t)} \in R^C$ represents the C-dimensional vector of variables at time $t$, and 
	\item $\boldsymbol{x} = \left[ \boldsymbol{x}_{(0)}, \boldsymbol{x}_{(1)}, ..., \boldsymbol{x}_{(C)}\right]^T$ \footnote{Here $T$ represents matrix transpose.} where $\boldsymbol{x}_{(c)} \in R^T$ represents the T-dimensional vector of signal from channel $c$.
\end{itemize}

Then the problem of anomaly detection on multivariate time series is to find a binary indicator $y\in \{0, 1\}$ representing whether the input signal is normal or abnormal.


\section {WaveletAE Architecture}

\begin{figure*}[!htb]
	\centering
	\includegraphics[width=0.88\textwidth]{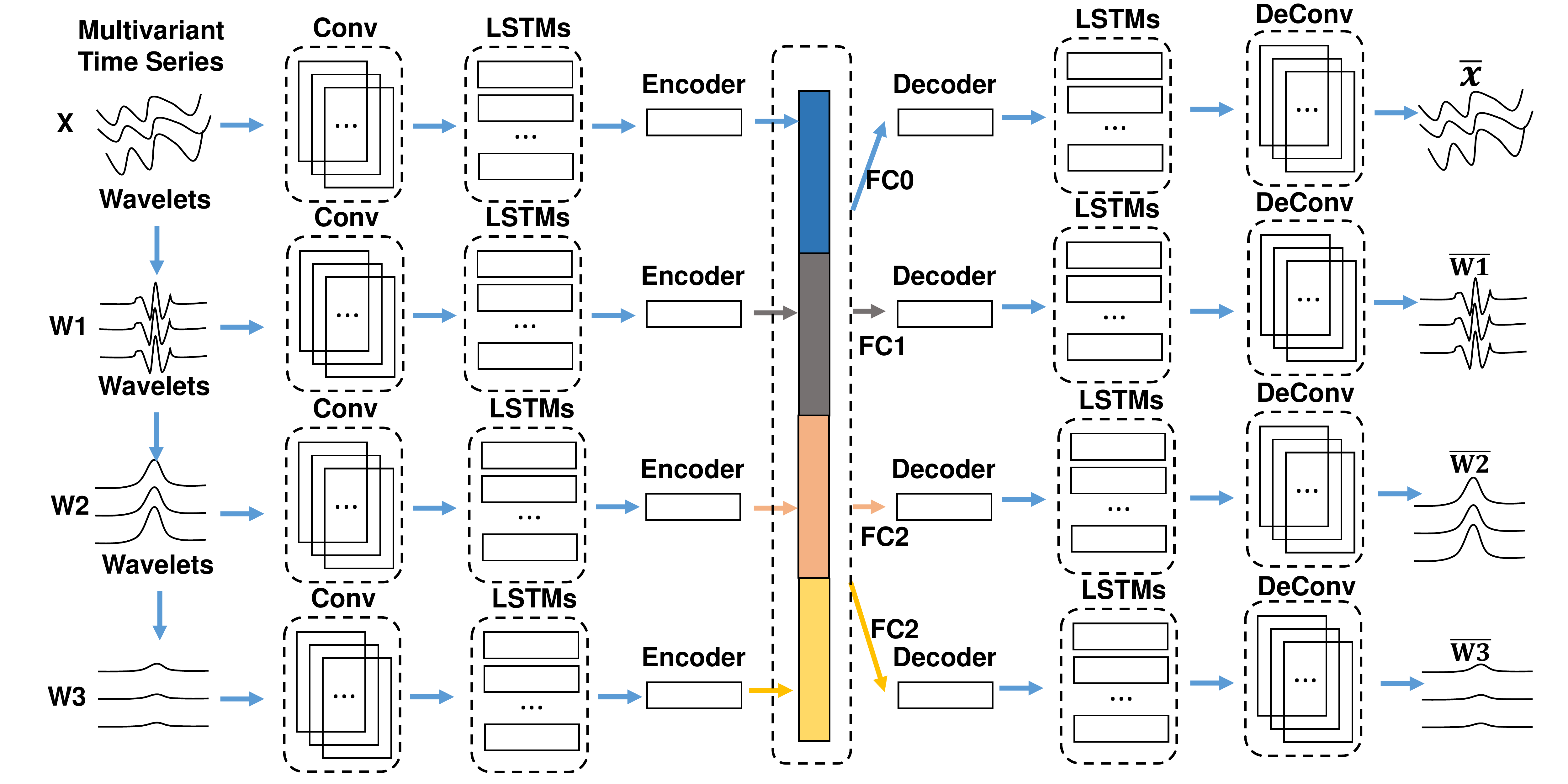}
	\caption{The architecture of WaveletAE. The input multivariate signals are first decomposed to multilevel wavelet detail coefficients. Then in each scale level, the original signals or wavelet detail coefficients go through a convolutional encoder and an LSTM encoder, the hidden states of the LSTM encoder will be concatenated to create a global code. In the decoding phase, fully connected layers will first map the global code to initial hidden states in each scale, then an LSTM decoder and a convolutional decoder will reconstruct the original signals and the wavelet detail coefficients.}
	\label{figure:WaveletAE}
\end{figure*}

In this section, we introduce the design of the WaveletAE architecture that can map both the complex channel-wise correlations and the dynamic multi-scale temporal patterns to a compact low-dimensional code to overcome the challenges in wind turbine icing detection. 
The overall structure of WaveletAE is illustrated in Figure \ref{figure:WaveletAE}. WaveletAE contains a multilevel discrete wavelet decomposition module, a convolutional encoder, a multiple scale LSTM encoder-decoder, and a convolutional decoder. The details of each component are enumerated in the rest of this section. 

\subsection{Multilevel Discrete Wavelet Decomposition}

Multilevel discrete wavelet decomposition (MDWD) extracts multilevel time-frequency features from time series. The decomposition reveals the variance of the signal in multiple scales, and recent research shows the advantages to combine MDWD with deep neural networks \cite{wang2018multilevel, zhao2018forecasting}. 

According to the classic pyramid algorithm, we first apply the 1-dimensional discrete wavelet transform on each input signal channel $\boldsymbol{x}_{(c)}, \: c \in \lbrace 0, 1, ..., C \rbrace$ to compute the wavelet detail coefficients in each channel to a specific level $L$, where $L$ can be viewed as a hyper-parameter of WaveletAE. The input signal $\boldsymbol{x} = \left[ \boldsymbol{x}_{(0)}, \boldsymbol{x}_{(1)}, ..., \boldsymbol{x}_{(C)}\right]^T$ is first augmented with the wavelet detail coefficients to multiple scales. Formally, in scale level $l \in \lbrace1, ..., L\rbrace$, the wavelet details coefficients are noted as $\boldsymbol{w}[l] = \left[ \boldsymbol{w}_{(0)}[l], \boldsymbol{w}_{(1)}[l], ..., \boldsymbol{w}_{(C)}[l]\right]^T$. It is worth noting that the approximation coefficients is not considered by the encoder for two reasons: (i) the approximation coefficients represent some smoothed averages of the input signal, such knowledge should be easily learned by the convolutional layers when processing the original signal; (ii) unlike the detail coefficients, approximation coefficients are not orthogonal to each other, the redundancy of the input can provide limited helpful information while enlarge the parameter space of the model.

\subsection{Convolutional Encoder}

As we illustrate in Figure \ref{figure:WaveletAE}, the original input signal along with the wavelet details in each scale are taken by independent convolutional layers. Assume the input for each convolutional encoder is $\boldsymbol{x} \in R^ {C\times T}$ or $\boldsymbol{w}[l] \in R^ {C\times \frac{T}{2^l}}$ noted as $\boldsymbol{a_{in}}$, the output of the convolutional layer $\boldsymbol{a_{out}}$ is defined as:
\begin{equation}
\boldsymbol{a_{out}} = f \left( \mathit{W} \ast \boldsymbol{a_{in}} + \mathit{b} \right),
\end{equation} 
where $f$ is activation function and $\ast$ denotes the convolutional operation, $\mathit{W}$ and $\mathit{b}$ are the parameters to learn in the convolutional encoder. Note that we set the kernel size of the first convolutional layer to the total channel number $C$, which forces the convolutional encoder to combine all the channels to capture the global correlations among channels.

\subsection{Multiple Scale LSTM Encoder-Decoder}

The activations generated from the convolutional encoder are then put to the next LSTM encoder independently for each scale. Following the definition of LSTM \cite{sak2014long}, we reform the definition in Formula \ref{lstm} to be consistent with the notation in Section \ref{preliminary_3}.

\begin{equation}
\label{lstm}
\begin{array}{r}
{\boldsymbol{i}^{(t)} =\sigma\left(W_{ii}\boldsymbol{a}^{(t)}+b_{ii} +W_{hi}\boldsymbol{h}^{(t-1)} + b_{hi} \right)}\\
{\boldsymbol{f}^{(t)} =\sigma\left(W_{if}\boldsymbol{a}^{(t)}+b_{if} +W_{hf}\boldsymbol{h}^{(t-1)} + b_{hf} \right)}\\
{\boldsymbol{g}^{(t)} =\tanh\left(W_{ig}\boldsymbol{a}^{(t)}+b_{ig} +W_{hg}\boldsymbol{h}^{(t-1)} + b_{hg} \right)}\\
{\boldsymbol{o}^{(t)} =\sigma\left(W_{io}\boldsymbol{a}^{(t)}+b_{io} +W_{ho}\boldsymbol{h}^{(t-1)} + b_{ho} \right)}\\
{\boldsymbol{c}^{(t)}=\boldsymbol{f}^{(t)} \cdot \boldsymbol{c}^{(t-1)}+\boldsymbol{i}^{(t)} \cdot \boldsymbol{g}^{(t)}}\\
{\boldsymbol{h}^{(t)}=\boldsymbol{o}^{(t)} \cdot \tanh \left( \boldsymbol{c}^{(t)}\right)}
\end{array}
\end{equation}

In the encoding phase, each LSTM encoder moves along the signal from timestamp $0$ to $\frac{T}{2^l}$ in scale level $l$. The final hidden states $\boldsymbol{h}[l]^{(\frac{T}{2^l})}$, where $l = {0, 1, 2, ...,  L}$, are concatenated to generate the global hidden state. Note that $l = 0$ represents the abstracted hidden states from the original input signals, while the $l = 1, ..., L$ denotes the hidden states from the decomposed multiple scales. 
In the decoding phase, each scale first uses an independent fully connected layer to map the global hidden state to initial hidden states for each LSTM decoder. Then the LSTM decoder moves along the signal in reverse order from timestamp $\frac{T}{2^l}$ to $0$ in the scale level $l$, while applies a fully connected layer to reconstruct the signal simultaneously. During training, the LSTM decoder in scale level $l$ uses the input from the convolutional encoder $\boldsymbol{a}[l]^{(t)}$ as input to obtain the hidden state $\boldsymbol{h}[l]^{(t-1)} $, while during inference, the predicted LSTM decoder output value at timestamp $t$ is input to the LSTM decoder for prediction at timestamp $t-1$. 

\subsection{Convolutional Decoder}

To reconstruct the original signal along with the multi-scale details, we need to decode the output sequence from the LSTM decoder. As a symmetric operation to convolutional layers, we employ the deconvolution operation formalized below:

\begin{equation}
\boldsymbol{a_{out}} = f \left( \mathit{W} \circledast \boldsymbol{a_{in}} + \mathit{b} \right)
\end{equation} 

where $f$ is the activation function the same as the convolutional encoder and  $\circledast$ denotes the deconvolutional operation, $\mathit{W}$ and $\mathit{b}$ are the parameters to learn of the convolutional decoder. Note that the last deconvolutional layer does not include an activation function to reconstruct the original signal and wavelet details. 

\subsection{Reconstruction Loss}

Finally, based the introduction of WaveletAE, the reconstruction loss for training the WaveletAE model and generating prediction in semi-supervised learning scenario is defined as:

\begin{equation}
\label{formula_loss}
loss =\lVert x - \bar{x} \rVert_2 + \sum\limits_{l=1}^L\lVert w[l]-\bar{w}[l]\rVert_2
\end{equation}

\section{Empirical Study}
\label{evaluation}

In this section, we describe the empirical experiments we have conducted to evaluate the proposed WaveletAE model \footnote{All the source code and datasets will be released after the acceptance.}. The aim is to answer the following question:

\noindent
\textit{How WaveletAE performs for the frozen blade detection task in the real world SCADA dataset from wind farms?}

To answer this question, we begin by presenting the dataset we collected from the real-world wind farms and reviewing the popular metrics for anomaly detection; then we compare the generalization performance of WaveletAE with the state-of-the-art approaches in both semi-supervised and supervised settings; finally, we discuss some practical tricks to generate robust predictions from WaveletAE under a simulated deployment.

\subsection{Experimental Setup}
\subsubsection{SCADA Dataset}
The data is provided by Goldwind Inc, one of the world's largest wind turbine manufacturers.
The raw data is collected from the supervisory control and data acquisition (SCADA) system which collects the monitoring data every $7$ second from hundreds of pre-installed standard sensors. According to the engineers' domain-specific knowledge, $26$ continuous variables relevant to frozen blades are preserved as the input multivariate signals.

Three wind turbines' monitoring data are obtained, which represents the running time of Machine $1$ for $306$ hours, Machine $2$ for $695$ hours and Machine $3$ for $329$ hours. 
We split the dataset as below:
\begin{itemize} 
	\item $60\%$ of the signal as the training set;
	\item $20\%$ of the signal as the validation set to compare the generalization performance between WaveletAE and the state-of-the-art approaches;
	\item $20\%$ of the signal as the final test data in the case study of simulated deployment.  
\end{itemize}

Further, the engineers from Goldwind Inc labeled the ranges, during which the blade icing occurs. For the training set and the validation, we split the raw signals into a collection of fragments of the fixed-length $512$ timestamps, approximately representing the wind turbine's status within one hour. Each fragment is associated with a binary label indicating whether the blades are frozen or not for this period. In the simulated deployment, we apply WaveletAE to generate accurate and robust detection of the blade icing situations. 

\subsubsection{Metrics for anomaly detection}
In the evaluation, we adopt a few general metrics for anomaly detection as briefly reviewed below: 
\begin{itemize}
	\item \textbf{Accuracy} is the number of correct predictions made by the model over all the predictions. 
	\item \textbf{Precision} is a measure that tells us what proportion of positions that we diagnose as an anomaly, actually are anomalies.
	\item \textbf{Recall} measures what proportion of samples that are anomalies is diagnosed by the model as an anomaly.
	\item \textbf{F1 score} is the Harmonic mean of precision and recall as a general evaluation of the model.
\end{itemize}

\subsection{Semi-Supervised Anomaly Detection}

Semi-supervised anomaly detection is required to construct models representing normal pattern from a given training data set which only include normal samples, and then output the likelihood of whether a test instance is normal or abnormal. The semi-supervised setting is popular in time series anomaly detection since in real-world applications, abnormal samples are difficult to obtain. We compare WaveletAE with the LSTM encoder-decoder model \cite{malhotra2016lstm}, which is one of the robust available state-of-the-art approaches for semi-supervised time series anomaly detection.

Under this setup, we include 671 normal fragments in the training set to train both WaveletAE and LSTM encoder-decoder \cite{malhotra2016lstm} for 16 epochs by Adam Optimizer \cite{kingma2014adam} with a learning rate of $0.001$. In the test phase, LSTM encoder-decoder applies the $L_2$ norm as the reconstruction error, while WaveletAE applies the loss function defined in Formula \ref{formula_loss}. 
The threshold $\tau$ for reporting the anomaly is defined as:

\begin{equation}
\tau = \beta \cdot mean\lbrace loss_{train} \rbrace
\end{equation}

where $\beta \in [1, 2]$ is a hyper-parameter, we set $\beta$ to $1.5$ in practice. 

The validation set includes $308$ fragments where $169$  is normal while $139$ is abnormal. The experimental results are listed in Table \ref{table_semi_supervised}.

\begin{table}[!htb]
	\centering
	\begin{tabular}{|c|c|c|}
		\hline
		Measurement & WaveletAE  & LSTM \cite{malhotra2016lstm} \\
		\hline
		Accuracy     & \textbf{0.493} &  0.428 \\
		Precision     & \textbf{0.471} &  0.437 \\ 
		Recall       & \textbf{0.986} &  0.928 \\
		F1 score      & \textbf{0.637} &  0.594 \\
		\hline
	\end{tabular}
	\caption{Comparison between WaveletAE and LSTM encoder-decoder\cite{malhotra2016lstm} under semi-supervised settings.}
	\label{table_semi_supervised}
\end{table}

\subsection{Supervised Anomaly Detection}

In our SCADA dataset, the training sets also include some range labels that indicate the segmentation is in an abnormal state. However, one risk of directly applying this collection of the fragment for training is that the dataset will be strongly biased according to the labeled ranges since the turbines function properly most of the time. To address this issue, we augment the number of positive samples (samples representing anomaly) by generating overlapping abnormal fragments and normal fragments without overlap regions. As we illustrated, the input multivariate time series is partitioned into a group of fragments of length 512, where each time series fragment approximately represents the wind turbine's status within one hour.  For example, suppose a wind turbine functions properly from $0:00$ to $8:00$, and there is a malfunction due to blading icing from $8:00$ to $10:00$, then we will cut the normal range without overlap so that 8 negative fragments each representing the state within on hour will be created, on the other hand, we can apply a step size of 10 minutes and a sliding window of one hour to move along the abnormal region and generate 7 positive fragments (Eg., $8:00$ to $9:00$, $8:10$ to $9:10$, etc.). In reality, we set the step size to the length of 16 (representing the signals of 112 seconds) to create a label balanced training dataset\footnote{Same method is applied to generate the validation sets for both semi-supervised and supervised settings.}.

In order to leverage the label information under supervised setting, we add a fully connected layer to map the global hidden states to the probability $p_a$ of the abnormality, the training loss becomes a affine combination of reconstruction loss and the cross-entropy binary classification loss $loss_{c} = -[y\log\left(p_a\right) + \left(1-y\right)\log\left(1- p_a\right)]$, defined as:

\begin{equation}
loss = \alpha \cdot loss_{re} + \left(1-\alpha \right)\cdot loss_{c}
\end{equation} 

where $\alpha$ is a hyper-parameter determining the weight between reconstruction loss and classification loss. For WaveletAE, the reconstruction loss is defined in Formula \ref{formula_loss}, while LSTM encoder-decoder still applies the $L_2$ norm as the reconstruction error with addtional binary classification loss. It is worth mentioning that a similar framework combining reconstruction loss and classification loss has been shown advantages in emotion classification \cite{khosla2018emotionx}. Besides LSTM encoder-decoder, we also include FCNN \cite{wang2017time}, which serves as a good baseline for time series classification.

Under the supervised setup, we include $671$ normal fragments and $550$ abnormal fragments in the training set to train WaveletAE, LSTM encoder-decoder \cite{malhotra2016lstm} and FCNN \cite{wang2017time} for $11$ epochs by Adam Optimizer \cite{kingma2014adam} with a learning rate of $0.001$. In the test phase, the classifiers' prediction is used to determine if the fragment is abnormal or not.  The experimental results are listed in Table \ref{table_supervised}.

\begin{table}[!htb]
	\centering
	\begin{tabular}{|c|c|c|c|}
		\hline
		Measurement & WaveletAE  & LSTM \cite{malhotra2016lstm}  & FCNN \cite{wang2017time}\\
		\hline
		Accuracy     &  \textbf{0.873}  & 0.792   & 0.747 \\
		Precision     &  \textbf{0.857}  & 0.804   & \textbf{0.857} \\
		Recall       &  \textbf{0.863}  & 0.712   & 0.525 \\
		F1 score      &  \textbf{0.861}  & 0.755   & 0.651 \\
		\hline
	\end{tabular}
	\caption{Comparison between WaveletAE and LSTM encoder-decoder\cite{malhotra2016lstm}, FCNN\cite{wang2017time} under supervised settings.}
	\label{table_supervised}
\end{table}

\subsection{Case Study: Simulated Deployment} 

Although WaveletAE shows advantages over state-of-the-arts anomaly detection algorithms on the validation set, it is worth discussing how the architecture can be deployed to generate robust real-time detection. To this end, we employ a sliding window vote schema. We first define two variables: a window size $T_w$ and a step size $T_s$, where ${T_s} \ll {T_w}$ and ${T_s}\left| {{T_w}} \right.$. Imagine that the time series is partitioned into blocks of length $T_s$, the schema lets an active window of length $T_w$ move along the input time series by a step of size $T_s$. The pre-trained WaveletAE will provide a prediction for the sequence within the active window. Each time the active window moves, a prediction will be made, so that all the blocks except the first block in the last active window will get a new prediction. In this way, each block will accumulate $T_w/T_s$ predictions as the sliding window moves along the signal. As a result, we can use a majority vote to determine if the current block is an anomaly or not. In our design, the majority vote can be even more flexible by setting a threshold $\tau$, if the ratio of the positive predictions is larger than or equal to the threshold, we will generate a positive prediction, otherwise a negative one.

To evaluate the performance of this schema, we use the labeled continues time series (unseen during training and validation) to simulate a real-time setting. The signal is then split into small blocks of length $T_s = 16$, to keep simplicity for each small block, we use a single label indicating the state of anomaly or not. The label is determined by whether half or more than half of the block falls into the anomaly regions. As the simulation begins, we accumulate blocks into the sliding window. Once the blocks fill up the sliding window, WaveletAE will make a prediction for the signal within the current window. When the next block arrives, the sliding window will move one step forward, so that the earliest arrived block will be abandoned and the latest arrived block will be placed at the end of the sliding window, then WaveletAE will make another prediction based on the newly updated time series within the sliding window. This simulation attempts to mimic the scenario, where the monitoring center fetches a signal block every $112$ seconds (consistent with $T_s = 16$) and combines this block with the previous blocks to make a prediction, the predicting results will be cached for majority vote; the voting results will indicate whether a blade icing situation is detected. In the above simulation, we set $T_w = 512$ the same as the fragment length in the training set.

We investigate the relationship between the threshold $\tau$ and the evaluation measurement and record the results in Table \ref{table_blade_vote_threshold}.   

\begin{table}[!htb]
	\centering
	\begin{tabular}{|c|c|c|c|c|}
		\hline
		$\tau$ & Accuracy  & Precision & Recall  & F1 score \\
		\hline
		0.1  &  0.865  &  0.569  &  \textbf{1.0}  &  0.725 \\
		0.2  &  0.910  &  0.665  &  \textbf{1.0}  &  0.799 \\
		0.3  &  0.955  &  0.799  &  \textbf{1.0}  &  0.888 \\
		0.4  &  \textbf{0.970}  &  0.895  & 0.943  &  \textbf{0.918} \\
		0.5  &  0.966  &  0.963  &  0.844  &  0.899 \\
		0.6  &  0.952  &  \textbf{0.989}  &  0.741  &  0.847\\
		0.7  &  0.925  &  0.986  &  0.588  &  0.736 \\
		0.8  &  0.907  &  0.983  &  0.488  &  0.652 \\
		0.9  &  0.891  &  0.979  &  0.394  &  0.562 \\
		\hline
	\end{tabular}
	\caption{Relationship between the voting threshold $\tau$ and the evaluation measurements.}
	\label{table_blade_vote_threshold}
\end{table}

\subsection{Discussion}
In both semi-supervised and supervised settings, we find that WaveletAE outperforms the state-of-the-art approaches in terms of accuracy, precision, recall, and $f_1$ score in Table \ref{table_semi_supervised} and Table \ref{table_supervised}, which verifies the design choice we make for WaveletAE to capture channel-wise dependences and dynamic temporal relationships in multiple scales. 

Unsurprisingly, WaveletAE generates more accurate predictions under the supervised learning setting than that under the semi-supervised setup. Besides the helpful information from the label in the training set, we believe that the loss function combining reconstruction error and classification error can also benefit the training procedure. Note that both WaveletAE and LSTM encoder-decoder (modified by us) that leverage such design, outperform the FCNN time series classification baseline in terms of accuracy, recall, and $f_1$ score. Intuitively, the reconstruction loss tends to force the hidden states to preserve more salient features of the input, so that noisy components of the time series will not influence the prediction, which improves generalization. Deep exploration of this phenomenon for other time-series analysis can lead to interesting independent work.

In the case study, we propose a schema to deploy the WaveletAE for real-time monitoring. Naturally, the majority vote will increase robustness for prediction since the minor inaccurate prediction will be corrected by the majority. Table \ref{table_blade_vote_threshold} enumerates the relationship between the voting threshold $\tau$ and the evaluation measurements. We can observe that $\tau$ determines the balance between precision and recall. In general, when $\tau$ increases, the precision increases while the recall decreases. This hyper parameter introduces flexibility for real-world scenarios. For example, engineers can choose small $\tau$ for a conservative strategy, when the cost for the de-icing procedure is relatively low while the damage of ice accumulation is extremely severe. In practice, $\tau$ can also be tuned dynamically. It is worth mentioning that in the simulation, the final prediction is made after the block accumulates all the votes, which may lead to latency in practice. However, this can be easily resolved by providing incremental monitoring to the control center --- once the first prediction is made, preliminary predictions can be reported; as the votes accumulate, the predictions become more accurate.

\section{Related Work}
\label{Related Work}

\subsection{Wind Turbine Prognosis via Applied Physics}


Wind farms usually locate in remote mountainous or rough sea regions, which makes monitoring and prognosis very challenging. A fault detection system will help to avoid premature breakdown, reduce maintenance cost, and to support for further development of a wind turbine \cite{ciang2008structural, lu2009review}. 
Traditionally, wind turbine prognosis researches from applied physics and mechanical engineering communities are focusing on design new physical detectors. Various detectors with advanced techniques have been proposed \cite{homola2006ice, munoz2018wavelet, munoz2016ice}.  
For example, \cite{homola2006ice} lists the detectors for icing accumulation, such as sensors damping ultrasonic waves on the wing surface, monitoring the resonant frequency of a probe, detecting piezoelectric pressure to estimate ice and turbulence, etc. However, these sensors' utility has some fundamental limitations: i) most of the sensors are effective only when placed on the blade instead of the nacelle, while associating the sensors to the flexing material of the blades are difficult; ii) the exposed cable connecting the sensors increases the risks of lightning for the wind turbine; iii) it is usually difficult to access the sensor in the event of failure.    

Another general approach for wind turbine icing detection is to design physical models by inferring ice accretion from power output signals, however, such approaches are usually validated by some software simulations without testing in the real world scenarios \cite{saleh2012wavelet}. 

\subsection{Data Driven Wind Turbine Prognosis}
Data-driven approaches for wind turbine state monitoring and failure detection have gained more attention due to the easy deployment comparing to installing complicated detectors. The method in \cite{regan2016wind} utilizes logistic regression and support vector machine (SVM) to conduct acoustics-based damage detection of wind turbines' enclosed cavities;
\cite{kuo1995intelligent}  and \cite{kusiak2012analyzing} apply neural networks to identify bearing faults and the existence of an unbalanced blade or a loose blade in wind turbines; 
\cite{malik2015application} implements a 3-layer probabilistic neural network (PNN) to diagnose wind turbine's imbalance fault identification based on generator signals;
\cite{zhang2012fault} includes both time domain and frequency domain information and tries variant classifiers to detect changes in the gearbox vibration excitement. 
Recently, \cite{chen2018learning} also studies blades' icing detection of wind turbines, where a feature representation of the signal is learned by clustering, and then used by k-nearest neighbors for prognosis. However, such nonparametric learning algorithms usually suffer seriously from overfitting, and the inference is more compute-intensive, which makes it impractical to deploy on real-time systems.

\subsection{Autoencoder Based Anomaly Detection}
In the data mining community, anomaly detection is an important problem that has been researched within various application domains \cite{chandola2009anomaly}. Based on the availability of the labeled dataset, anomaly detection techniques fall into three main categories: a supervised mode where the labeled instances for both normal and anomaly classes are accessible, the semi-supervised mode where the training set only includes normal samples and unsupervised mode that does not require any training data. Recently, generative models, especially autoencoders \cite{kieu2019outlier, malhotra2016lstm, su2019robust, wang2019self, zhang2019deep} 
have drawn increasing attention for time series anomaly detection, where a representation of the time series in a compact space is learned by the encoder to reconstruct the original signals by the decoder. According to \cite{vskvara2018generative}, the generative model shows promising potential and should be studied in greater depth. Unfortunately, none of these proposed approaches have overcome all the challenges we meet in the wind turbine icing detection problem, where both complex channel-wise correlation and dynamic multi-scale temporal patterns have to be captured in the model.


\section{Conclusion}
\label{conclusion}
We present WaveletAE, a novel autoencoder architecture to resolve the difficulties in monitoring multivariate signals for wind turbine's blade icing detection. WaveletAE can simultaneously learn the multiple channel correlations and multiple scale dynamic patterns in analyzing multivariate time series data. The generalization performance of WaveletAE has been verified under both semi-supervised and supervised anomaly detection settings in real-world SCADA dataset. A case study of simulated deployment demonstrates the robustness and flexibility of WaveletAE for real-time monitoring. 

\bibliographystyle{plain}
\bibliography{WaveletAE}

\end{document}